\documentclass{article}

\PassOptionsToPackage{numbers}{natbib}
\usepackage[preprint]{neurips_2021}

\usepackage[utf8]{inputenc} % allow utf-8 input
\usepackage[T1]{fontenc}    % use 8-bit T1 fonts
\usepackage{hyperref}       % hyperlinks
\usepackage{url}            % simple URL typesetting
\usepackage{booktabs}       % professional-quality tables
\usepackage{amsfonts}       % blackboard math symbols
\usepackage{nicefrac}       % compact symbols for 1/2, etc.
\usepackage{microtype}      % microtypography
\usepackage{xcolor}         % colors

\usepackage{mathtools}
\usepackage{cleveref}

\usepackage{my_macros}

\title{Preprocessing Reward Functions for Interpretability}

\author{%
  Erik Jenner\thanks{Work done at the Center for Human-Compatible AI} \\
  University of Amsterdam\\
  \texttt{erik@ejenner.com} \\
  \And
  Adam Gleave \\
  UC Berkeley \\
  \texttt{gleave@berkeley.edu}
}

\newcommand{\gridworldcaption}{Each heatmap shows the rewards for all possible transitions in a \(10 \times 10\) gridworld. The circle in the center of each square represents the reward for staying in that state. The four triangles in each square represent the reward of transitions \emph{leaving} that square in each of the four directions.}

\newcommand{\mountaincarcaption}{Each plot shows the reward during a rollout over five episodes (separated by the gray vertical lines).}

\begin{document}

\maketitle

\begin{abstract}
  In many real-world applications, the reward function is too complex to be manually specified.
  In such cases, reward functions must instead be learned from human feedback.
  Since the learned reward may fail to represent user preferences, it is important to be able to validate the learned reward function prior to deployment.
  One promising approach is to apply interpretability tools to the reward function to spot potential deviations from the user's intention.
  Existing work has applied general-purpose interpretability tools to understand learned reward functions.
  We propose exploiting the intrinsic structure of reward functions
  by first \emph{preprocessing} them into simpler but equivalent reward functions,
  which are then visualized. 
  We introduce a general framework for such reward preprocessing and propose concrete preprocessing algorithms.
  Our empirical evaluation shows that preprocessed rewards are often significantly easier to understand than the original reward.
\end{abstract}

\section{Introduction}
Reinforcement learning (RL) agents have reached superhuman performance in many tasks, such as games, with clearly defined objectives~\citep{silver2016mastering,openai2019dota,vinyals2019grandmaster}.
However, real-world deployment of RL is often hampered by the difficulty of specifying an appropriate reward function by hand.
A variety of methods to learn reward functions have been developed to address this challenge.
These algorithms learn from human feedback such as demonstrations~\citep{ng:2000,ramachandran:2007,ziebart:2008,fu:2018,bahdanau:2019,ibarz:2018,brown:2019},
preferences~\citep{akrour:2011,wilson:2012,christiano:2017,sadigh:2017,ziegler:2019,ibarz:2018,brown:2019},
or even the initial state of the environment~\citep{shah2018preferences}.

This gives rise to a new problem: how can we evaluate the learned reward function, and spot potential failure modes before deployment?
We could train a policy on the reward model, and then evaluate this policy, such as by humans judging rollouts from the policy.
However, this approach has serious drawbacks. 
Training a policy can be very expensive in complex environments, which makes it hard to quickly compare different learned rewards.
It is also brittle: if the trained policy doesn't perform well, it is unclear whether the fault lies with the reward function or the policy training procedure.

Moreover, we might want to know how well a learned reward \emph{transfers} to different environment dynamics.
Indeed, transfer is a major motivation for learning a reward rather than just directly learning a policy.
But it is often challenging to specify the set of dynamics under which we would like the reward to be robust.
Even where this is possible, training and evaluating a policy on all such environments is typically infeasible.

To address these problems, prior work introduced the EPIC distance~\citep{gleave2021quantifying} to quantify the difference between two reward functions.
When a ground truth reward is available, we can evaluate a learned reward simply by computing
its EPIC distance to the ground truth.
However, reward learning is most useful precisely when we do \emph{not} have access to a ground truth reward.
In this setting, EPIC may still have some limited utility for comparing and perhaps clustering several learned reward models, but it cannot tell us if any of these learned models are correct.

\Citet{michaud2020understanding} instead suggest \emph{interpreting} reward models to verify they capture user preferences.
They find existing interpretability methods such as saliency maps can help understand reward models.
However, they also found significant limitations in this approach, concluding that ``reward interpretability may need significantly different methods from policy interpretability''.

We believe that significant advances in reward interpretability can be made by taking advantage of the special structure of reward functions.
In particular, many different reward functions are \emph{equivalent}, in the sense that they induce the same optimal policies -- no matter the environment dynamics.
Given a learned reward model, we can apply transformations that do not change the optimal policy, but simplify the reward function.
We can then visualize this simplified reward instead of the original.
We call this approach \emph{reward preprocessing}, as we ``preprocess'' the reward model prior to visualization.

Our framework for reward preprocessing involves two key components:
1) a class of reward transformations that yield equivalent reward functions in some sense (e.g.\ by preserving the optimal policy under arbitrary environment dynamics), and
2) an objective that measures how interpretable a given reward function is.
We then optimize over the class of transformations using the given objective
to find the most interpretable equivalent reward function.
A key property of this framework is that the learned reward model is treated as a black box.
This means that it may use an arbitrary function approximator and can be learned using any reward learning algorithm and feedback modality.

In summary, our key contributions are: 
1) a novel framework that exploits the intrinsic structure of reward functions
    to increase their interpretability before visualization.
2) Two concrete applications of this framework, using different objectives for interpretability.
3) An empirical evaluation, finding that our methods often significantly improve interpretability.

\section{The Reward Preprocessing Framework}
Our interpretability method operates on a reward function \(r(s, a, s')\), where
\(s\) is the current state, \(a\) is the action taken in that state, and \(s'\) is the
next state.
Our method only requires the ability to evaluate \(r\): there are no restrictions on how \(r\) is computed or how it was learned.
From \(r\), we produce a simpler but equivalent reward function \(r'\), which we then visualize. 

Concrete instantiations of this framework must make two choices.
First, they must specify which reward functions are deemed equivalent via an equivalence relation \(\sim\).
Second, they must provide some measure of ``simplicity'' or ``interpretability'', represented by a cost function \(J\).
We then seek to find a minimum cost reward function \(r'\) that is equivalent to \(r\):
\begin{equation}\label{eq:general_optimization_problem}
r' := \argmin_{\hat{r} \sim r} J(\hat{r}).
\end{equation}

In the following, we discuss how to choose the equivalence relation \(\sim\) and cost function \(J\).

\subsection{Equivalence Relation}

We would like to treat two rewards as equivalent if they will produce the same behaviour in the intended downstream application.
It is known that potential shaping~\citep{Ng1999policyinvariance} and rescaling by a positive constant never change the ordering of policies.
It is therefore safe to treat such rewards as equivalent for most applications.

However, some applications permit a broader notion of equivalence.
For example, if the reward model will only ever be used for policy optimization in a specific task, then we can include any transformations that preserve optimal policies in that task.
A simple example is \(S'\)-redistribution: moving reward between different successor states, while preserving \(\mathbb{E}_{S'} r(s,a,S') = \mathbb{E}_{S'} r'(s,a,S')\).
This will not change the optimal policy, so long as the transition dynamics determining $S'$ remain fixed.
\citet{skalse2021invariance} characterize a variety of such equivalence classes under varying assumptions.

\subsection{Choosing Cost Functions}
The cost function \(J\) should represent the interpretability of a reward function.
Of course, no simple objective can capture the entire concept of interpretability since reward functions may be interpretable for a variety of reasons.
For example, \emph{sparse} rewards are often interpretable, as the user can pay attention only to the few transitions on which the agent receives a non-zero reward.
However, a dense reward could still be easy to understand if it has some other simple structure: for example, taking on only two different values depending on which region of the world the agent is in.

Instead of looking for a single cost function that completely characterizes interpretability,
we therefore suggest using \emph{multiple} cost functions, each of which describes
some condition that is \emph{sufficient} but not \emph{necessary} for interpretability.
We can then find an optimal equivalent reward for each of the cost functions and present all of these rewards for the user to choose between. 
We can also rank the reward functions provided the cost functions are on a comparable scale, presenting lowest-cost rewards first.

Another factor determining the appropriate cost functions is the method used for visualization.
For example, a reward function that has sparse output is ideal if we wish to show the user the reward of particular transitions.
However, we might prefer sparsity in the \emph{features} that the reward depends on if using higher-level visualization methods like saliency maps.

\section{Methodology}

In this section, we describe a few simple concrete instances of our reward preprocessing framework.
Despite their simplicity, we find in \cref{sec:results} that they nonetheless can yield significant improvements.
However, these choices are likely far from optimal, and so should be viewed as establishing a lower bound on the benefit obtainable from reward preprocessing.

\subsection{Potential Shaping Equivalence Relation}

We define two rewards to be equivalent, \(r \sim r'\), if they are equal up to \emph{potential shaping}~\citep{Ng1999policyinvariance}.
Specifically, \(r \sim r'\) if there exists some real-valued state-only function \(\Phi\) called a \emph{potential} for which \(r'(s, a, s') = r(s, a, s') + \gamma\Phi(s') - \Phi(s)\,\), where \(\gamma \in [0, 1)\).
Potential shaping changes the returns of an episode by only the potential \(\Phi(s_0)\)
of the initial state (in the finite horizon case, the potential of terminal states is restricted to be zero). Since the policy does not affect the initial state, the ordering
over policies is invariant under potential shaping. 
This holds for \emph{arbitrary} transition dynamics and initial state distributions.
Therefore, rewards related to each other by potential shaping can be considered equivalent even under transfer
to different environment dynamics.

A notable advantage of potential shaping for our purposes is that it is very easy to optimize over the resulting equivalence class.
We simply parametrize the potential as a neural network \(\Phi_\theta(s)\) with parameters \(\theta\). 
Then the optimization problem from \cref{eq:general_optimization_problem} becomes
\begin{equation}\label{eq:potential_shaping_optimization}
  \argmin_{\theta} J(r'_\theta)\,,\:\text{where}\:r'_\theta(s, a, s') = r(s, a, s') + \gamma\Phi_\theta(s') - \Phi_\theta(s).
\end{equation}

We optimize \cref{eq:potential_shaping_optimization} using (stochastic) gradient descent.
This requires a differentiable cost function \(J\), but does \emph{not} require the reward function \(r\) to be differentiable.

Rewards differing by a positive scale factor also produce the same policy ordering.
However, since our visualization techniques can handle rewards at a range of scales, we choose to preserve the scale during preprocessing.
Accordingly, we do not include rescaled rewards as equivalent.

\subsection{Cost Functions}

We evaluate two types of cost functions: a sparsity-inducing one based on the \(L^1\) norm and a smoothness-inducing measure of absolute deviation.
In tabular (gridworld) settings, we evaluate these cost functions on a uniform distribution \(\mathcal{D}\) over all possible transitions.
In continous control environments, we evaluate on transitions sampled from the same distribution \(\mathcal{D}\) used for visualization.

Sparse rewards are easy to understand as the user only needs to attend to rewards with non-zero transitions.
However, the \(L^0\) norm is non-differentiable.
Moreover, even if the ground-truth reward is sparse, learned reward functions are usually not exactly equivalent to a sparse reward due to the presence of noise.
We therefore use two different relaxed notions of sparsity:
the \(L^1\) norm \(\abs{r}\) and the slightly transformed version \(\log(1 + \abs{r})\).
In particular, we minimize:
\begin{equation}\label{eq:sparsity_cost}
  J_{\text{sparse}}(r) := \mathbb{E}_{(s, a, s') \sim \mathcal{D}} \,f\big(r(s, a, s')\big)\,,
\end{equation}
where \(\mathcal{D}\) is the distribution over transitions and \(f(x)\) is either \(\abs{x}\) or \(\log(1 + \abs{x})\).

An alternative is to minimize the fluctuations between rewards of transitions adjacent in time.
This creates a smoothly varying reward signal.
The user can then understand the reward by looking at the trend over time.
This frees the user from having to attend to the reward at every single transition, similar to sparsity.
Again, we use an \(L^1\) and a logarithmic version of such a smoothness cost:
\begin{equation}\label{eq:smoothness_cost}
  J_{\text{smooth}}(r) := \mathbb{E}_{(s_t, a_t, s_{t+1}, a_{t+1}, s_{t+2}) \sim \mathcal{D}}
  \,f\big(r(s_t, a_t, s_{t+1}) - r(s_{t+1}, a_{t+1}, s_{t+2})\big)\,.
\end{equation}

\section{Results}\label{sec:results}

We evaluate our methods in two environments: a gridworld, with varying rewards, and the classic mountain car continuous control task~\citep{boyan1995generalization}.
We test our method with a mixture of hand-designed and learned rewards.
The hand-designed rewards consist of a simple ground-truth reward, with shaping and/or noise added to challenge the preprocessing method.
The learned rewards are trained via either adversarial inverse reinforcement learning~\citep[AIRL]{fu:2018}, or via deep reinforcement learning from human preferences~\citep[DRLHP]{christiano:2017}.
Both methods are trained on synthetic data, consisting of rollouts from an expert policy (AIRL) or preference comparisons induced by the ground-truth reward (DRLHP).

In gridworld experiments, we use a tabular potential and reward model.
That is, we learn a separate value \(\Phi(s)\) and \(r(s,a,s')\) for each state and transition.
In mountain car, we use small MLPs for the reward model and potentials, except for some cases where a linear potential is sufficient to find a simple equivalent reward.
Our code is available at \url{https://github.com/HumanCompatibleAI/reward-preprocessing}.

\subsection{Simplifying Shaped Rewards}
We start by testing our method in a gridworld setting~\citep{mazelab}.
While unrealistic, gridworlds have the considerable benefit of allowing the entire reward function to be easily visualized.
This therefore allows a more thorough evaluation of our method than in other tasks.

\begin{figure}
  \centering
  \includegraphics[width=\textwidth]{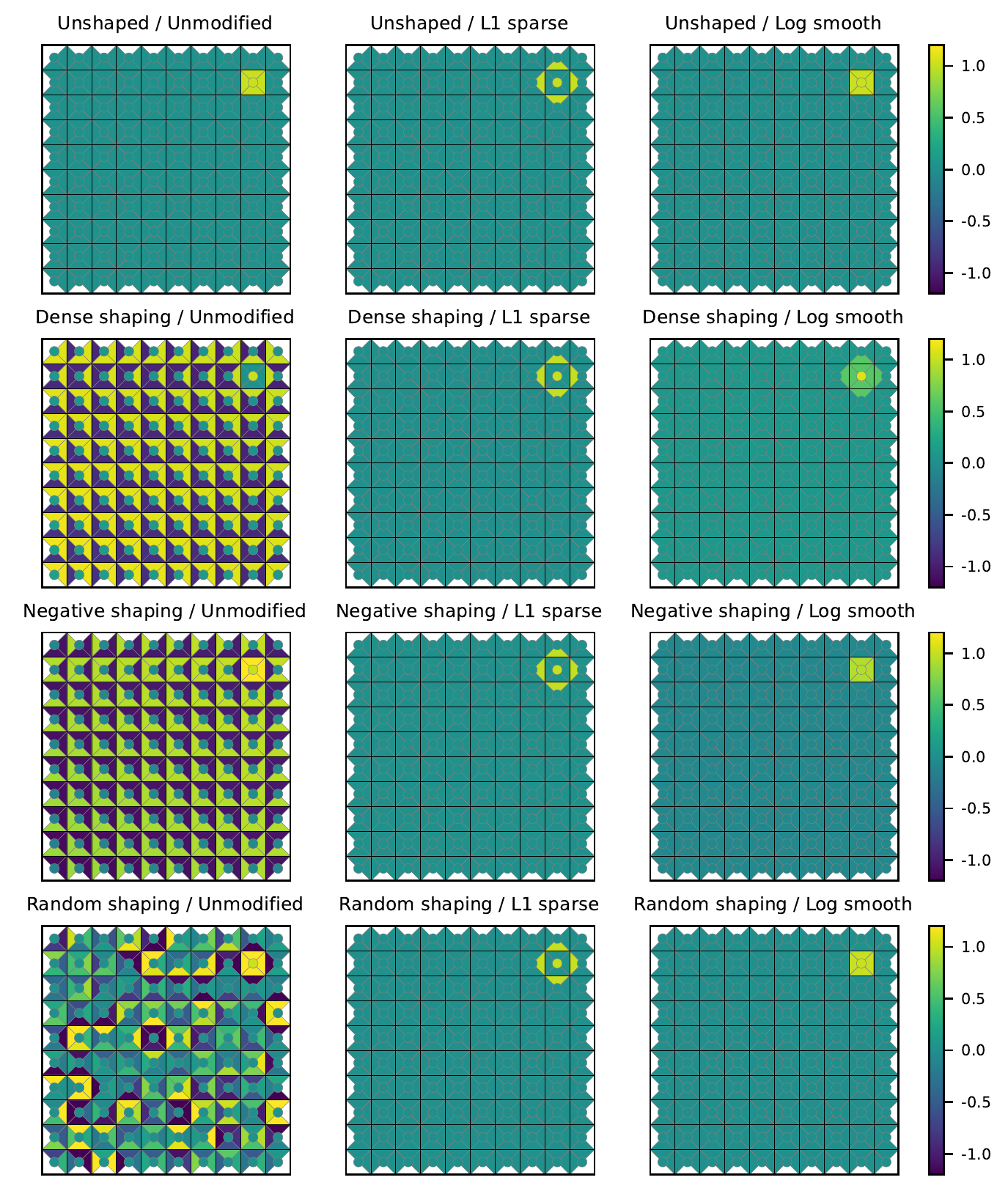}
  \caption{
    Preprocessing can recover a sparse reward from complex shaping.
    The original sparse \texttt{Goal} reward is shown in the top-left, with three shaped versions below.
    These rewards are shown after preprocessing with the sparsity (middle) and smoothness (right) cost functions.
    The preprocessed rewards are easy to understand, and are similar across a range of shaping.
    \gridworldcaption{}
  }
  \label{fig:gridworld_ground_truth_10_goal}
\end{figure}

\begin{figure}
  \centering
  \includegraphics[width=\textwidth]{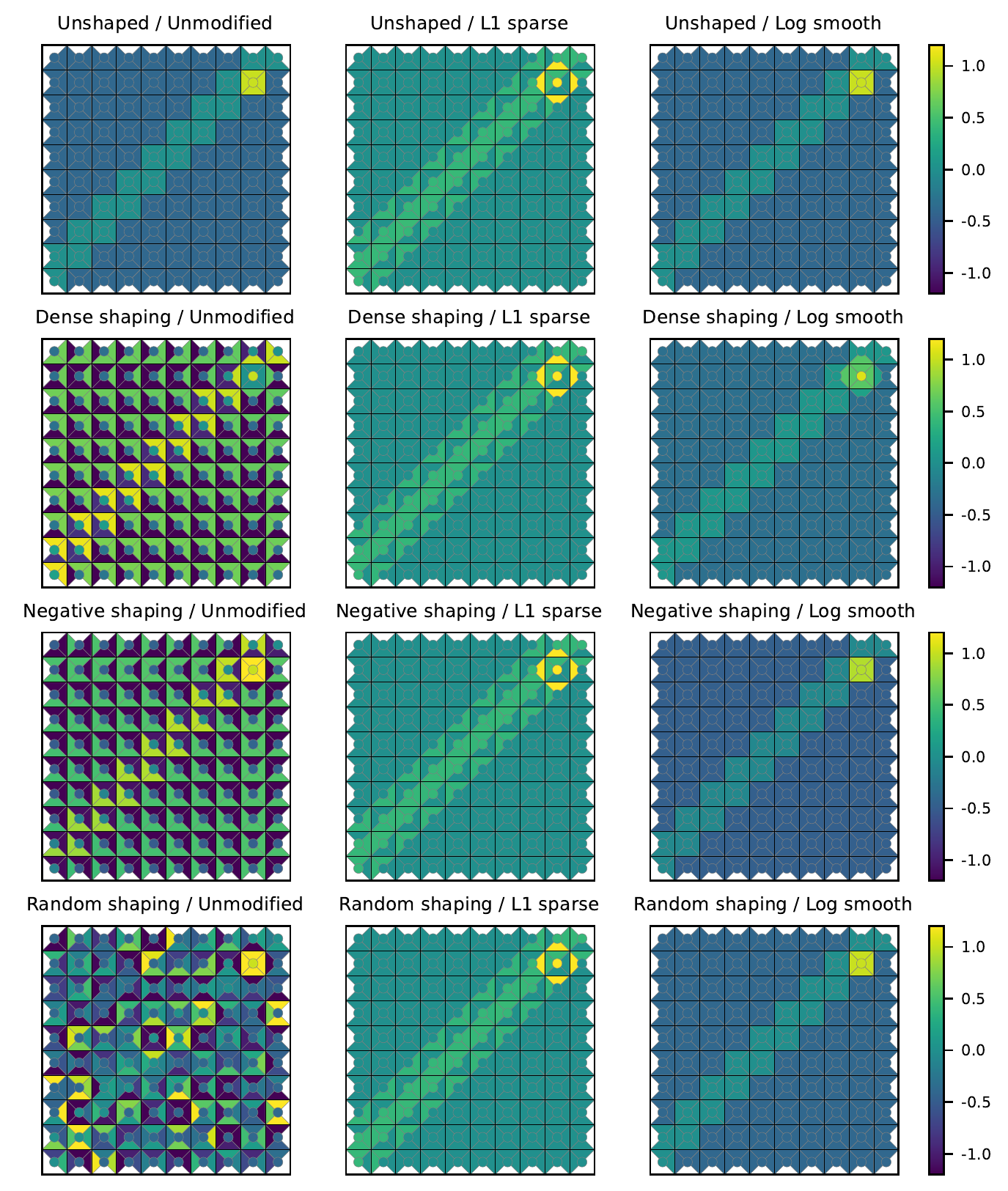}
  \caption{
    Preprocessing can recover a simpler dense reward from complex shaped rewards.
    The original \texttt{Path} reward (top-left) incentivizes following a diagonal path to the goal state.
    The three shaped versions below largely obscure this pattern, but preprocessing is able to recover something similar to the original reward.
    This is notable as the original reward is not sparse, so still achieves a relatively high cost under the \(L^1\) norm, but is nonetheless lower cost than the highly complex shaped rewards.
    \gridworldcaption{}
  }
  \label{fig:gridworld_ground_truth_10_path}
\end{figure}

In \cref{fig:gridworld_ground_truth_10_goal,fig:gridworld_ground_truth_10_path}, we visualize gridworld rewards before (leftmost column) and after (middle and right column) our preprocessing methods.
In the \texttt{Goal} environment in \cref{fig:gridworld_ground_truth_10_goal}, the reward is simply \(1\) on a single goal square in the top right corner and $0$ everywhere else.
This is readily understood and our preprocessing largely retains this reward unchanged.
However, when we add shaping with the Manhattan distance from the goal (second row), the reward becomes much harder to understand.
Our preprocessing, however, is able to simplify this shaped reward to something close to the original sparse objective.
Similar results hold for the negative Manhattan distance from the goal (third row) and the particularly confusing random shaping (last row).

In the \texttt{Path} environment in \cref{fig:gridworld_ground_truth_10_path}, the original reward (top left) prefers a specific path for reaching the goal state.
Once again, the shaped versions obscure this, but preprocessing reliably recovers a simple and interpretable reward.

In these plots, we use the \(L^1\) version of the sparsity cost and the logarithmic version of the smoothness cost.
These work slightly better than the other versions, but the difference is very small.
The results for all versions can be found in \cref{fig:gridworld_ground_truth_10_goal_l1,fig:gridworld_ground_truth_10_goal_log,fig:gridworld_ground_truth_10_path_l1,fig:gridworld_ground_truth_10_path_log} in the appendix.

\subsection{Understanding Learned Rewards}
In the previous experiment, all the reward functions were exactly equivalent to the simple original reward.
By contrast, learned reward models may be noisy or contain systematic errors, and may not be equivalent to any simple reward.
To evaluate how our method performs in this more realistic setting, we trained reward models from demonstrations (AIRL) and preference comparisons (DRLHP) on synthetic data in both of the previous \texttt{Goal} and \texttt{Path} environments.
The results of applying our preprocessing method are shown in \cref{fig:gridworld_drlhp_10_goal_l1,fig:gridworld_drlhp_10_goal_log,fig:gridworld_drlhp_10_path_l1,fig:gridworld_drlhp_10_path_log,fig:gridworld_airl_10_goal_l1,fig:gridworld_airl_10_goal_log,fig:gridworld_airl_10_path_l1,fig:gridworld_airl_10_path_log} in the appendix.

For the reward model learned using preference comparisons (DRLHP), even the preprocessed models look very noisy.
The goal state does tend to be somewhat more visible in the preprocessed than the unprocessed rewards, but neither are easy to understand.
The reward model learned by AIRL differs even more from the ground truth reward, and potential shaping is unable to bridge that gap.

It might be possible to remove more of the noise by using a larger equivalence class than potential shaping.
However, expanding the equivalence class might mean the preprocessed reward would no longer induce the same optimal policy as the unmodified reward in some environment dynamics.
Indeed, the fact that potential shaping is not sufficient to remove the noise suggests that what DRLHP and AIRL have learned is not just a complex but validly shaped version of the ground truth reward.


\subsection{Mountain Car}
\begin{figure}
  \centering
  \includegraphics[width=\textwidth]{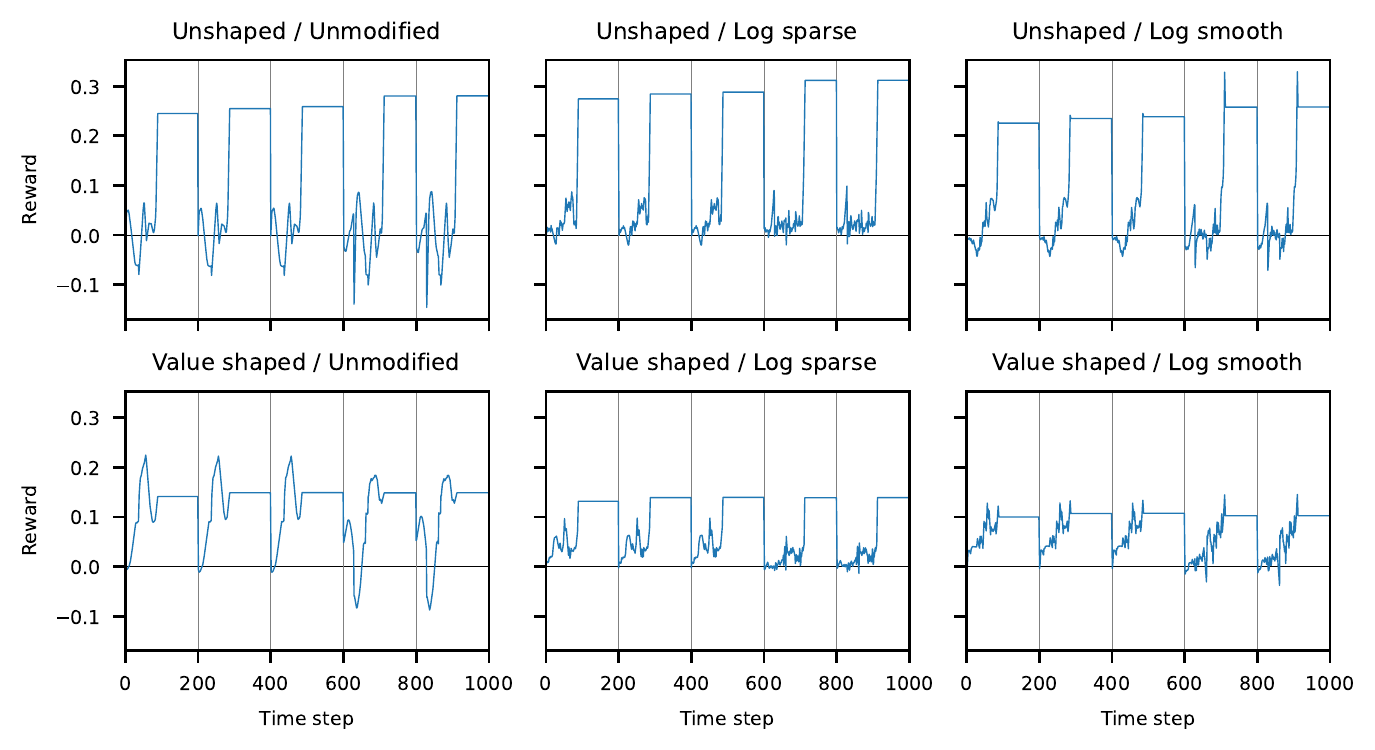}
  \caption{Preprocessing can simplify complex learned reward models for mountain car. The left column shows reward models learned using synthetic preference comparisons based on the ground-truth reward (top), and the ground-truth shaped with an optimal value function (bottom). Preprocessing for sparsity (middle) and smoothness (right) produces simpler and less noisy reward curves, especially in the shaped setting. \mountaincarcaption{}}
  \label{fig:mountain_car_drlhp_log}
\end{figure}

Since the mountain car environment has an infinite number of possible transitions, we cannot plot the rewards of all possible transitions as we did in the gridworld tasks.
Instead, we visualize reward functions by plotting the reward signal over time during expert trajectories.
\Cref{fig:mountain_car_drlhp_log} visualizes two learned reward models (left) and the reward signal after preprocessing with a log sparse (middle) and log smooth (right) cost function.

The model in the top row was trained using DRLHP on synthetically generated preferences. Specifically, we sampled Boltzmann-rational preferences between trajectory fragments based on the ground truth reward.
In the second row, we first learned an optimal state value function for the mountain car environment and then used this to shape the ground truth reward before generating preferences. 
This simulates human feedback, which may be shaped compared to a sparse ground truth since humans already reward incremental progress~\citep{christiano:2017}.

As in the gridworld setting, both the learned and preprocessed rewards are noisy.
However, the preprocessed reward functions are still significantly simpler than the learned models, especially in the shaped case.
The sparsity cost function performs better here than the smoothness cost. \Cref{fig:mountain_car_drlhp_log} uses the logarithmic version of both but the \(L^1\) version in \cref{fig:mountain_car_drlhp_l1} yields almost exactly the same results.

Notably, the residual noise after preprocessing in \cref{fig:mountain_car_drlhp_log} is likely not removable by potential shaping.
In particular, we find in \cref{fig:mountain_car_ground_truth,fig:mountain_car_ground_truth_l1} that preprocessing on shaped versions of the ground-truth reward recover simple, noise-free rewards.
The residual noise is therefore likely an accurate depiction of errors in the learned reward.

\section{Related Work}
Interpreting reward models has recently begun to receive some attention.
\Citet{russell19explaining} apply standard interpretatibility methods from supervised learning
to reward functions. Specifically, they use feature importance estimates from a simple fitted global model and from LIME~\citep{ribeiro2016lime} to interpret the reward function.

Globally fitting a simpler model to a reward function has some similarities to our reward preprocessing approach. 
However, a major difference is that the simple model will usually not be equivalent to the original reward function.
In contrast, we learn an \emph{equivalent} but still simplified reward model.
This is possible because we exploit the structure that reward functions naturally have, whereas \citet{russell19explaining} only apply preexisting interpretability methods.

\Citet{michaud2020understanding} also apply existing interpretability methods to understand reward models.
In contrast to Russell and Santos, they work directly with the given reward, without fitting a simpler model.
They suggest and combine three different approaches, namely gradient saliency maps, occlusion
maps and hand-crafted counterfactual inputs.
All of these methods can also be applied to supervised learning more broadly and do not take advantage of the structure of reward functions.

Our reward preprocessing framework is complementary to these methods for interpreting reward functions.
We advocate first preprocessing a given reward to select a maximally comprehensible equivalent reward function.
The resulting reward function can then be visualized or otherwise interpreted using a range of techniques.

While there is only a handful of work seeking to understand learned reward functions, considerably more work has focused on interpreting \emph{policies}~\citep{puiutta2020survey}.
One approach is to learn a policy from a class of intrinsically simple functions rather than neural networks~\citep{verma2018pirl}. 
Alternatively, \citet{juozapaitis2019decomposition} present a method that explains policy actions by an additive decomposition of $Q$-values.
Another promising recent direction is using causal models to explain policy behavior~\citep{madumal2020explainable,deletang2021causal}.
%Numerous other methods for policy interpretability have been developed, and are surveyed by \citet{puiutta2020survey}.

\Citet{devidze2021} approach interpretability of reward functions from a different angle: rather than interpreting a complex \emph{learned} reward function, they aim to \emph{design} a reward function that trades off between interpretability (operationalized as sparsity) and ease of policy optimization.

\section{Limitations and Future Work}
One limitation of our approach is that while potential shaping does not change the optimal policy, it can make the policy optimization problem easier or harder.
Consequently, the policy learned by an RL algorithm might well differ between the unmodified learned reward and the theoretically ``equivalent'' reward used for visualization.
This issue is most significant in environments where policy optimization can be challenging.
Reasoning about how shaping affects RL algorithm performance is challenging, so this is only a significant factor when the tool is being used by trained practitioners.

The above limitation is a way in which potential shaping can be too \emph{big} an equivalence class.
However, there is also a sense in which it is too \emph{small}.
In practice, we do not usually care about a reward function transferring to \emph{all} possible transition dynamics.
If it is known the transition dynamics satisfy certain invariants, then we may be able to use a larger equivalence class while still guaranteeing optimal policy preservation.

In addition to modifying the equivalence class, there are also numerous alternative cost functions that could be employed. In particular, the cost functions we suggest are targeted at the visualizations we use in this paper.
Other visualizations might benefit from different cost functions.
However, it seems likely that the basic concepts of sparsity and smoothness will be useful in many settings.
For example, visualizations using gradient saliency maps might benefit from maximizing the sparsity of the gradients, rather than of the rewards themselves.

\section{Conclusion}
We have introduced a novel framework to preprocess reward functions prior to visualization.
Our empirical results demonstrate this methodology can recover simple reward functions from shaped versions of ground-truth rewards.
Moreover, our method can substantially simplify even noisy learned reward models.
However, some low-quality learned reward models are still difficult to understand even with our method, suggesting that reward learning algorithms often converge to models significantly different from the user's intended preferences.

\bibliography{references}
%%%%%%%%%%%%%%%%%%%%%%%%%%%%%%%%%%%%%%%%%%%%%%%%%%%%%%%%%%%%

\appendix
\section{Appendix}

\begin{figure}
  \centering
  \includegraphics[width=\textwidth]{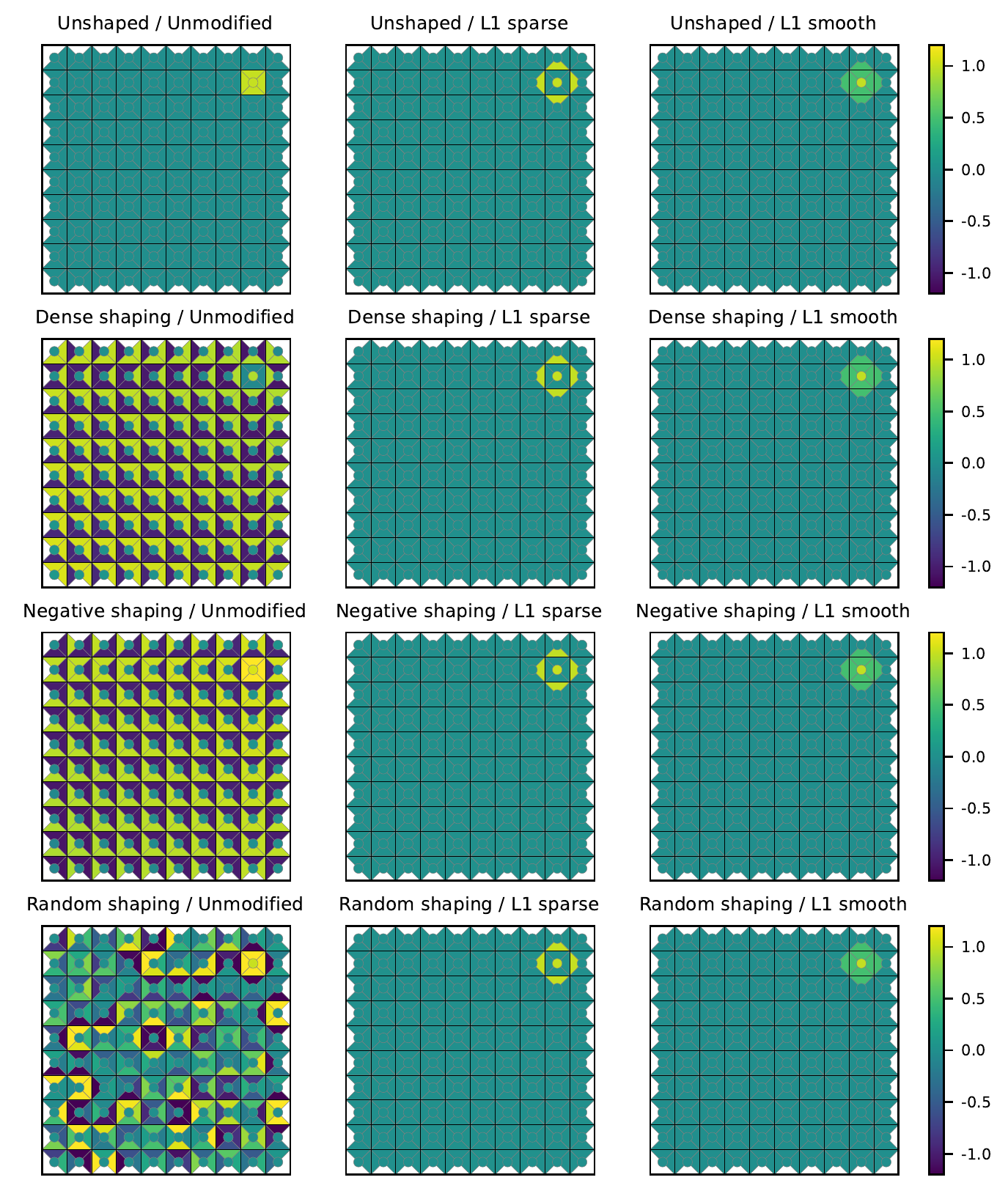}
  \caption{\texttt{Goal} reward preprocessed with \(L^1\) versions of the sparsity and smoothness cost function. The results are very similar to those in \cref{fig:gridworld_ground_truth_10_goal}.
    \gridworldcaption{}
  }
  \label{fig:gridworld_ground_truth_10_goal_l1}
\end{figure}
\begin{figure}
  \centering
  \includegraphics[width=\textwidth]{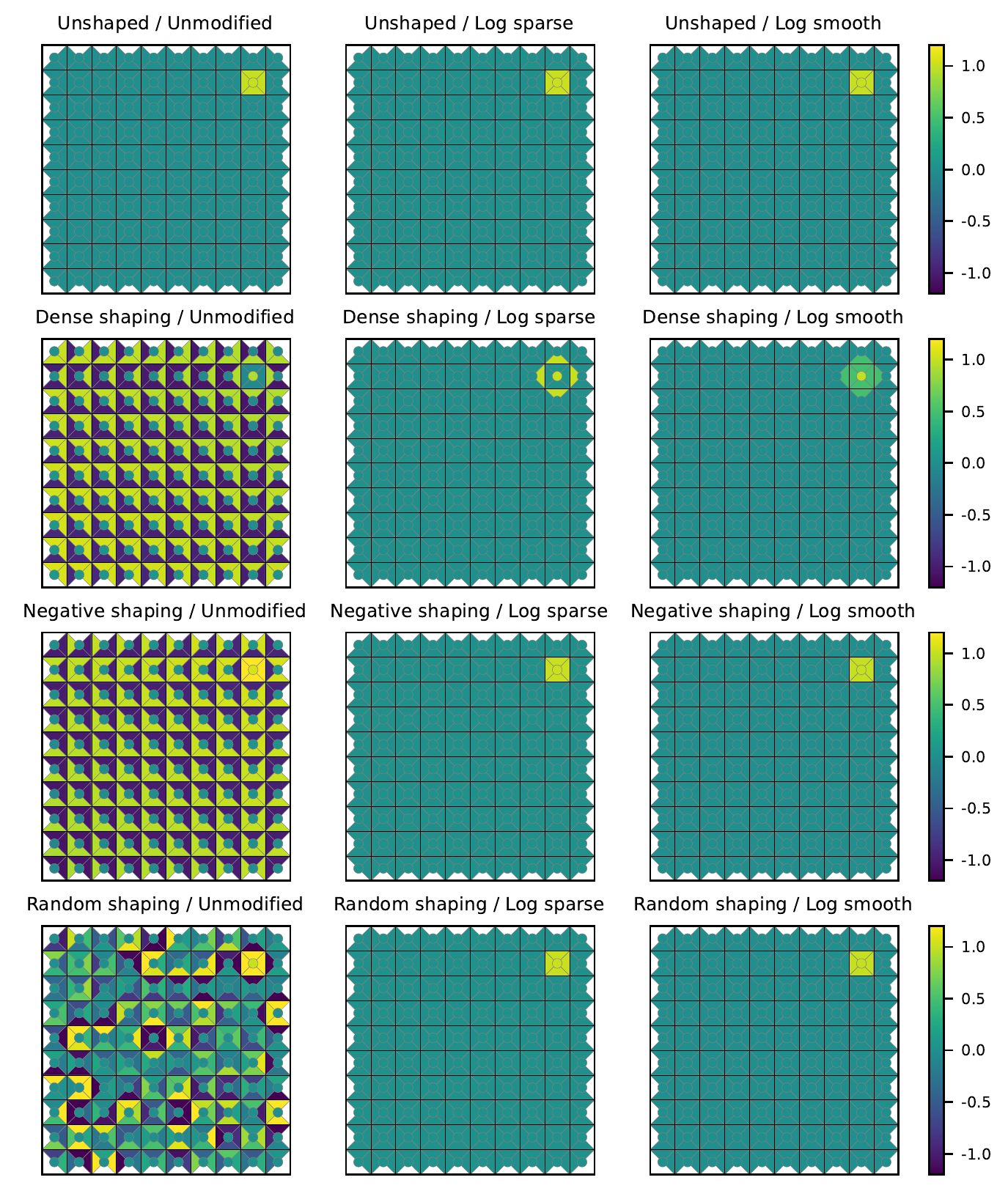}
  \caption{
    \texttt{Goal} reward preprocessed with logarithmic versions of the sparsity and smoothness cost functions. Again, the results are qualitatively similar to those in \cref{fig:gridworld_ground_truth_10_goal}.
    \gridworldcaption{}
  }
  \label{fig:gridworld_ground_truth_10_goal_log}
\end{figure}

\begin{figure}
  \centering
  \includegraphics[width=\textwidth]{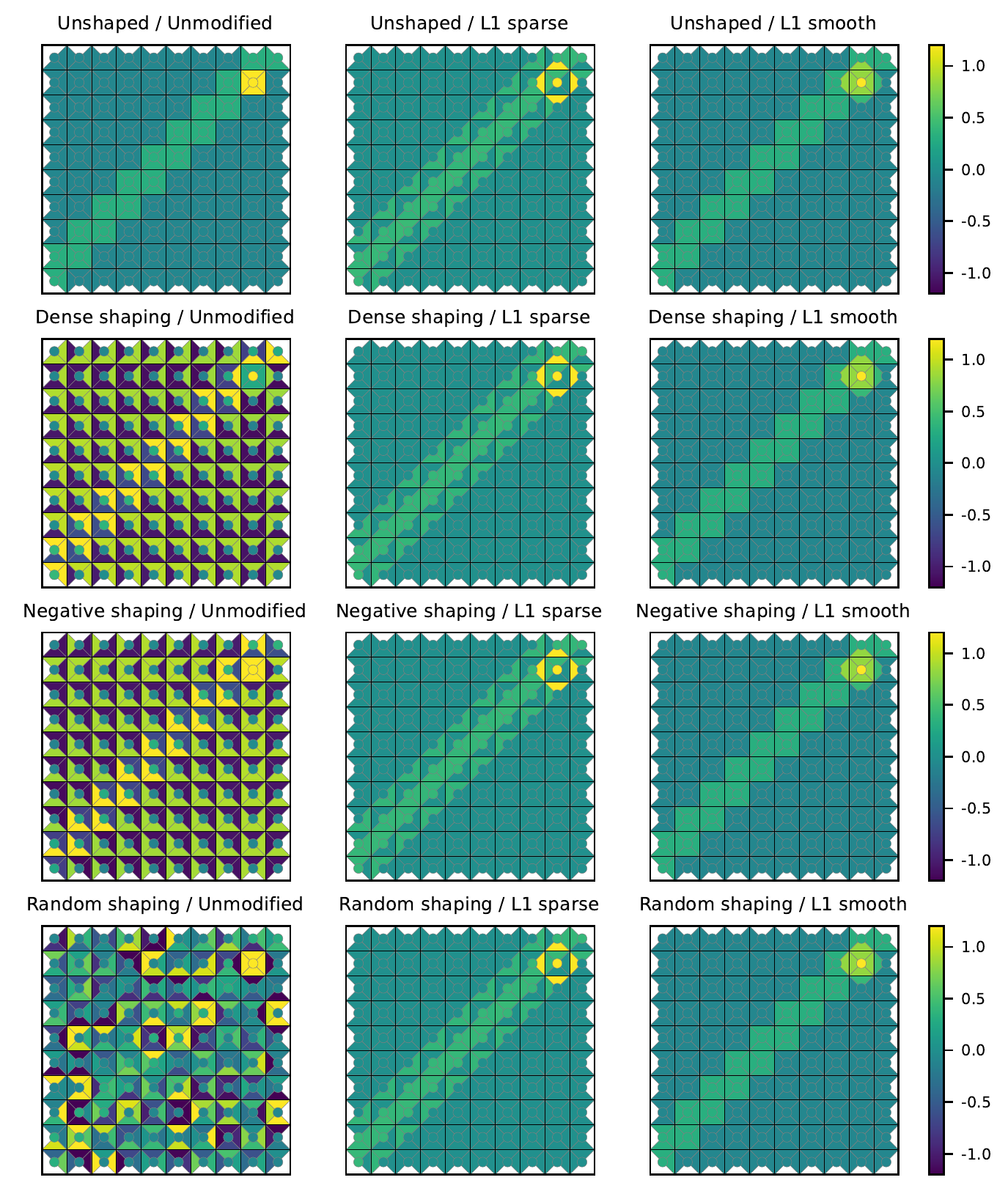}
  \caption{
    \texttt{Path} reward preprocessed with \(L^1\) versions of the sparsity and smoothness cost function. The results are very similar to those in \cref{fig:gridworld_ground_truth_10_path}.
    \gridworldcaption{}
  }
  \label{fig:gridworld_ground_truth_10_path_l1}
\end{figure}
\begin{figure}
  \centering
  \includegraphics[width=\textwidth]{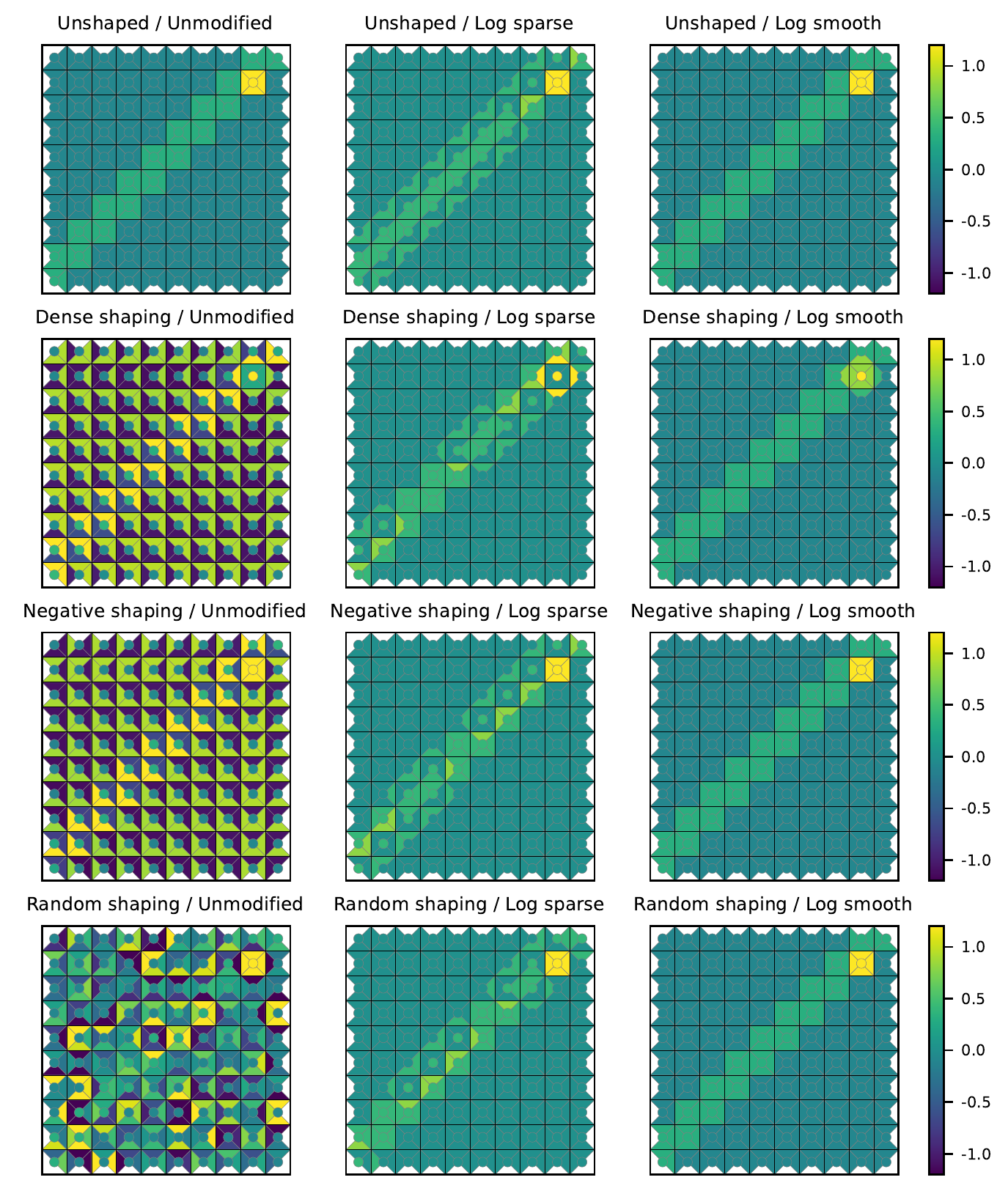}
  \caption{
    \texttt{Path} reward preprocessed with logarithmic versions of the sparsity and smoothness cost functions. Compared to the \(L^1\) sparse cost function in \cref{fig:gridworld_ground_truth_10_path_l1}, the log sparse cost recovers a slightly less symmetric but still significantly simplified reward.
    \gridworldcaption{}
  }
  \label{fig:gridworld_ground_truth_10_path_log}
\end{figure}

\begin{figure}
  \centering
  \includegraphics[width=\textwidth]{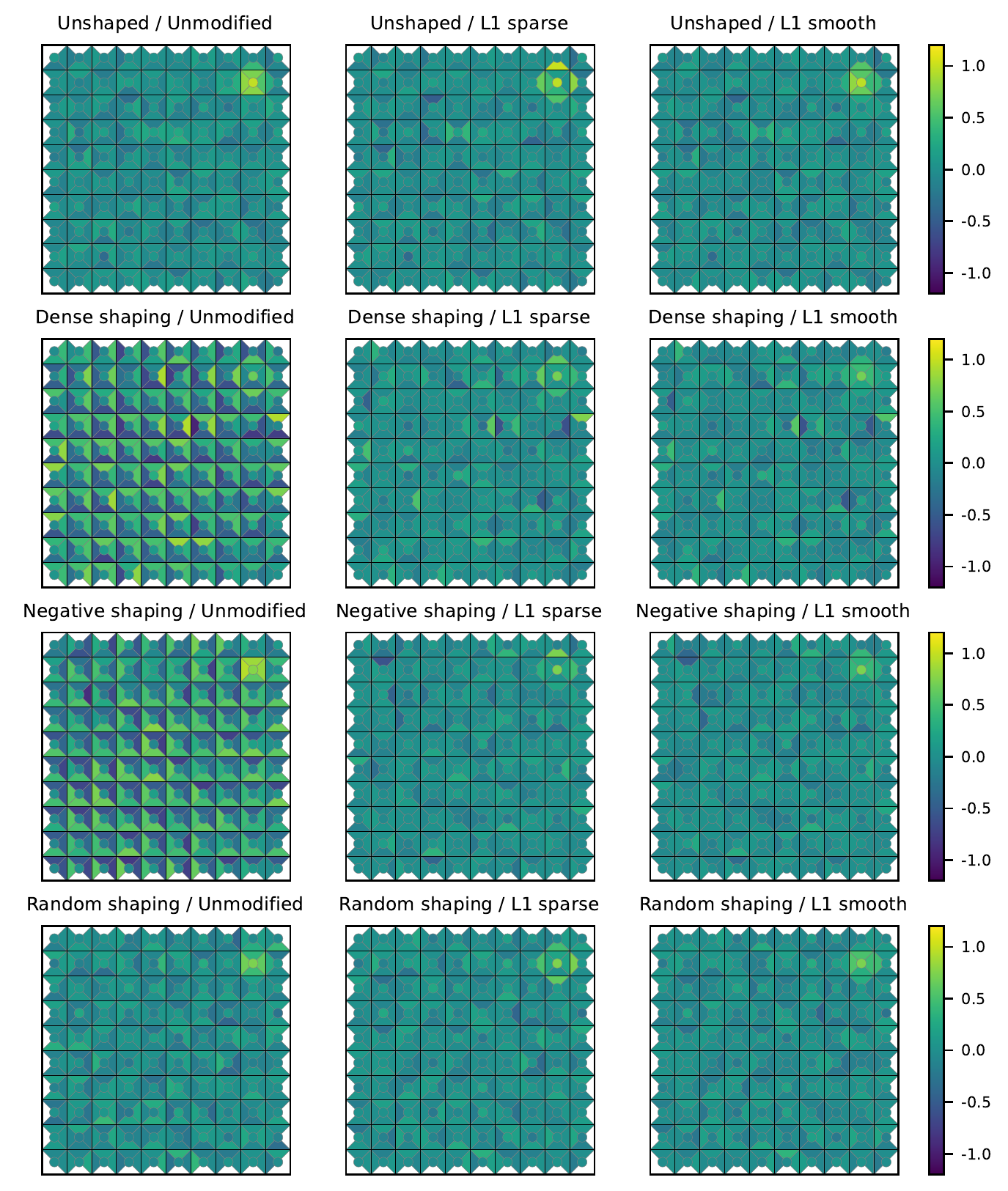}
  \caption{
    Reward models trained on synthetic data from the \texttt{Goal} reward using preference comparison (leftmost column) and preprocessed versions of these (middle and right). The cost functions used here are the \(L^1\) version of the sparsity and smoothness cost. \gridworldcaption{}
  }
  \label{fig:gridworld_drlhp_10_goal_l1}
\end{figure}

\begin{figure}
  \centering
  \includegraphics[width=\textwidth]{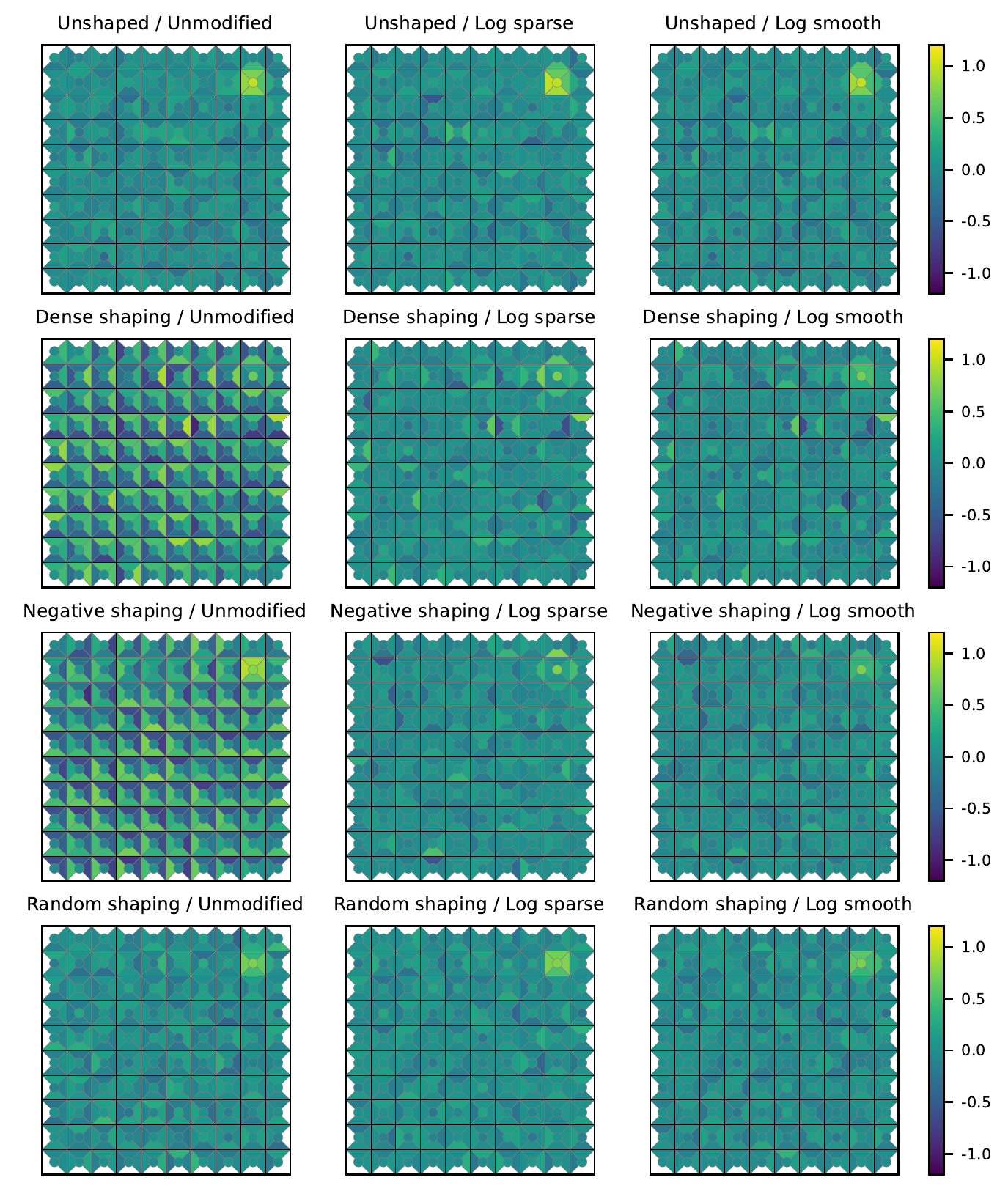}
  \caption{
    Reward models trained on synthetic data from the \texttt{Goal} reward using preference comparison (leftmost column) and preprocessed versions of these (middle and right). The cost functions used here are the logarithmic version of the sparsity and smoothness cost. \gridworldcaption{}
  }
  \label{fig:gridworld_drlhp_10_goal_log}
\end{figure}

\begin{figure}
  \centering
  \includegraphics[width=\textwidth]{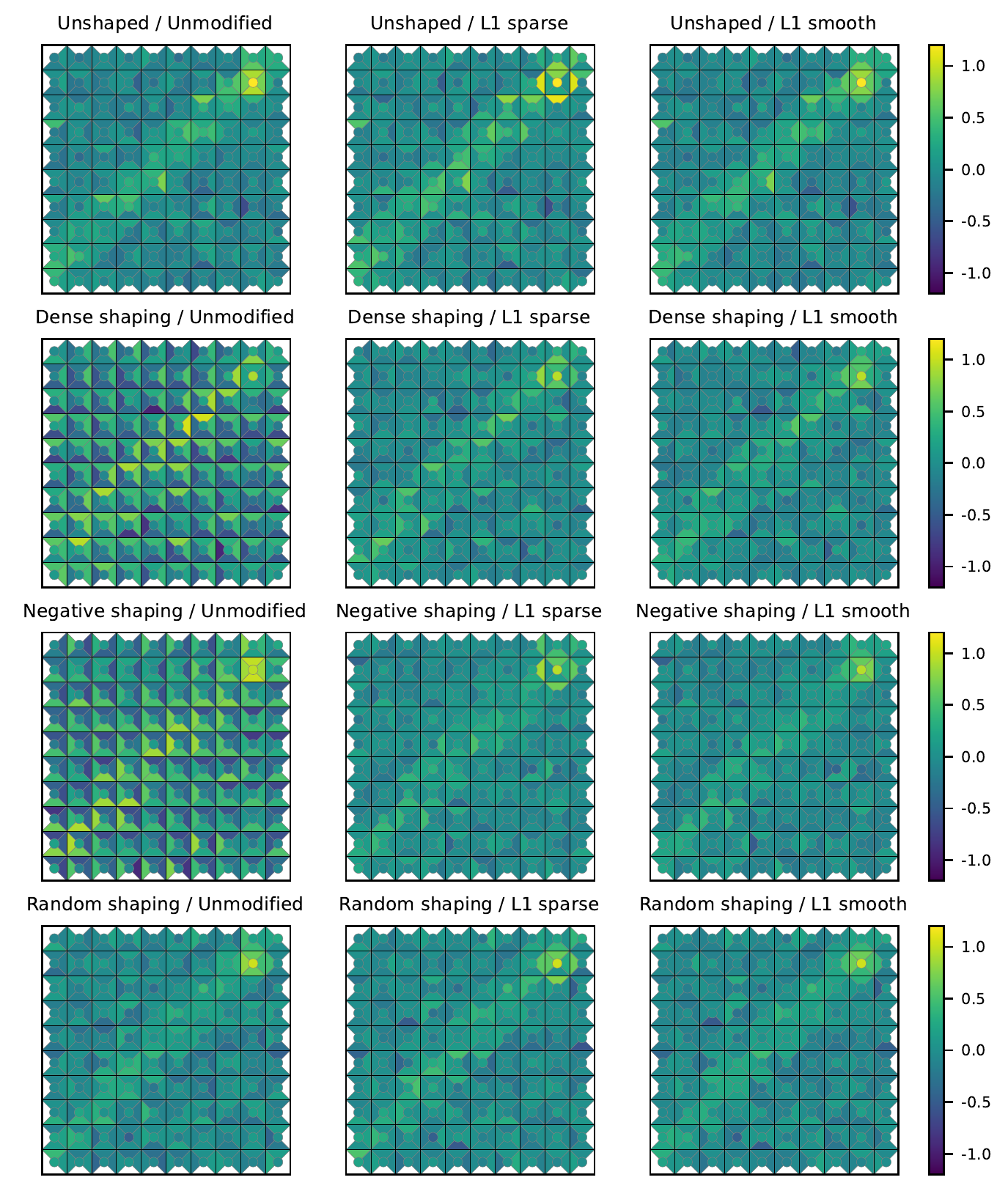}
  \caption{
    Reward models trained on synthetic data from the \texttt{Path} reward using preference comparison (leftmost column) and preprocessed versions of these (middle and right). The cost functions used here are the \(L^1\) version of the sparsity and smoothness cost. \gridworldcaption{}
  }
  \label{fig:gridworld_drlhp_10_path_l1}
\end{figure}

\begin{figure}
  \centering
  \includegraphics[width=\textwidth]{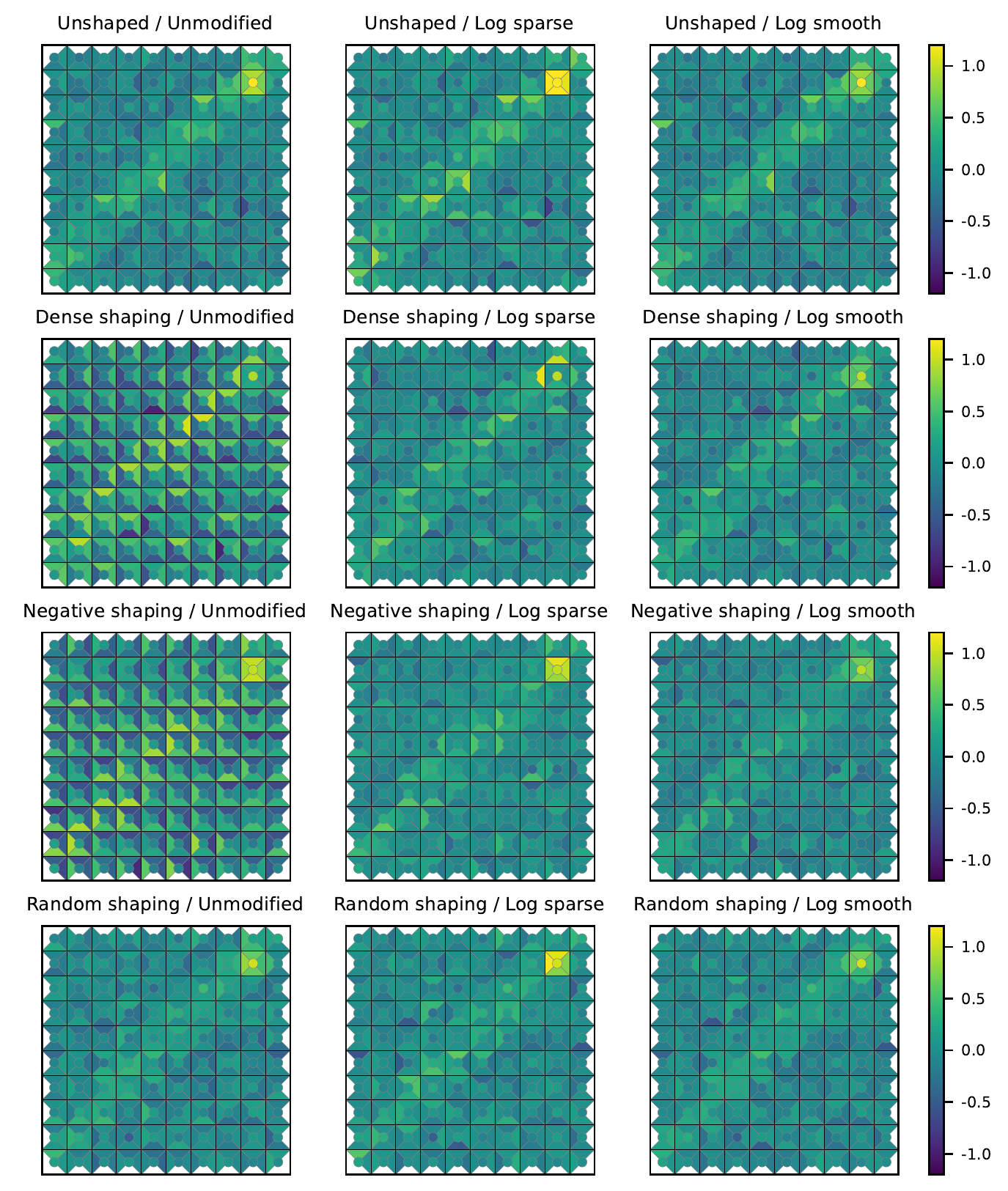}
  \caption{
    Reward models trained on synthetic data from the \texttt{Path} reward using preference comparison (leftmost column) and preprocessed versions of these (middle and right). The cost functions used here are the logarithmic version of the sparsity and smoothness cost. \gridworldcaption{}
  }
  \label{fig:gridworld_drlhp_10_path_log}
\end{figure}

\begin{figure}
  \centering
  \includegraphics[width=\textwidth]{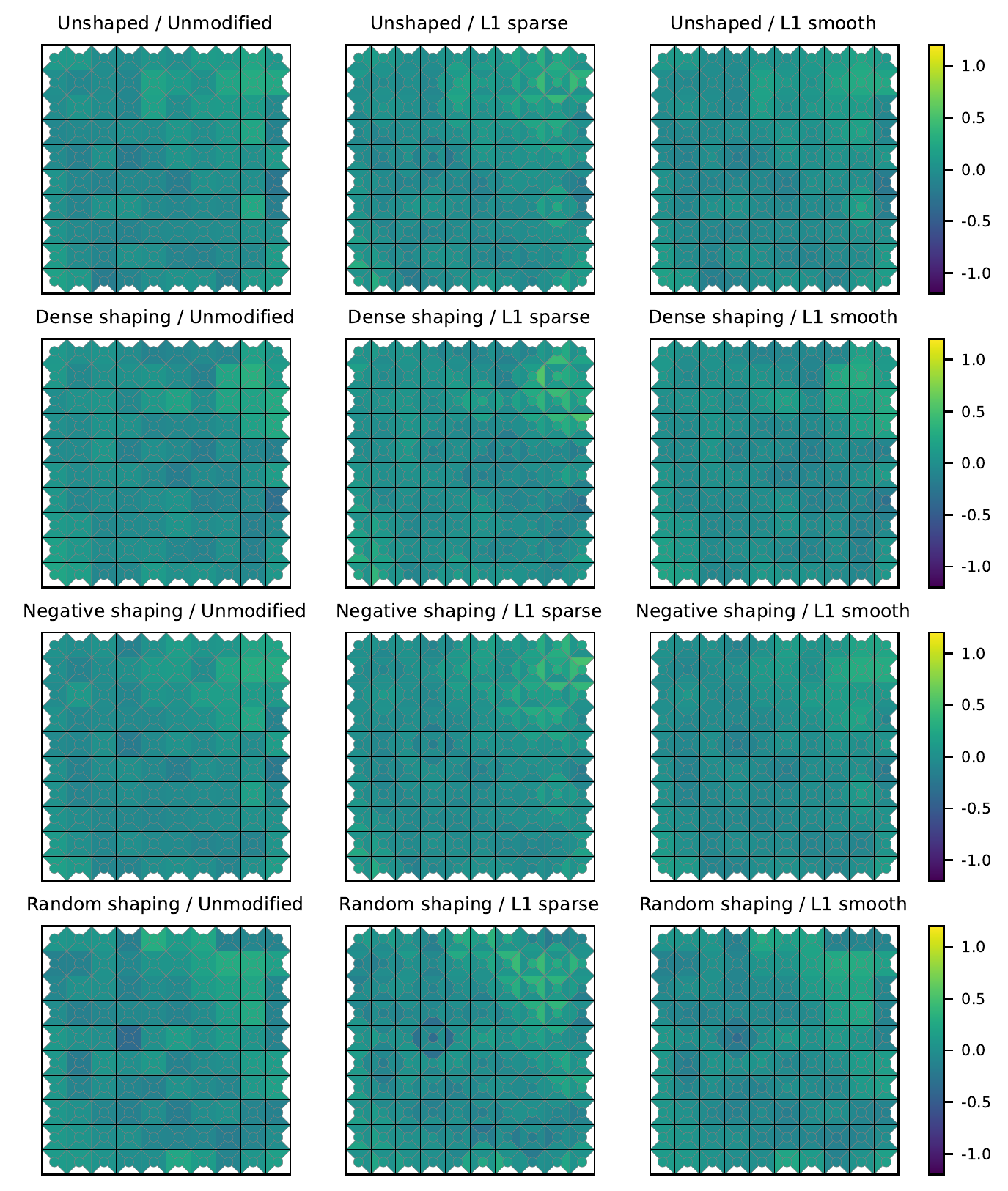}
  \caption{
    Reward models learned using AIRL from expert demonstrations for the \texttt{Goal} reward (leftmost column) and preprocessed versions of these (middle and right). The cost functions used here are the \(L^1\) versions of the sparsity and smoothness cost.
     \gridworldcaption{}
  }
  \label{fig:gridworld_airl_10_goal_l1}
\end{figure}
\begin{figure}
  \centering
  \includegraphics[width=\textwidth]{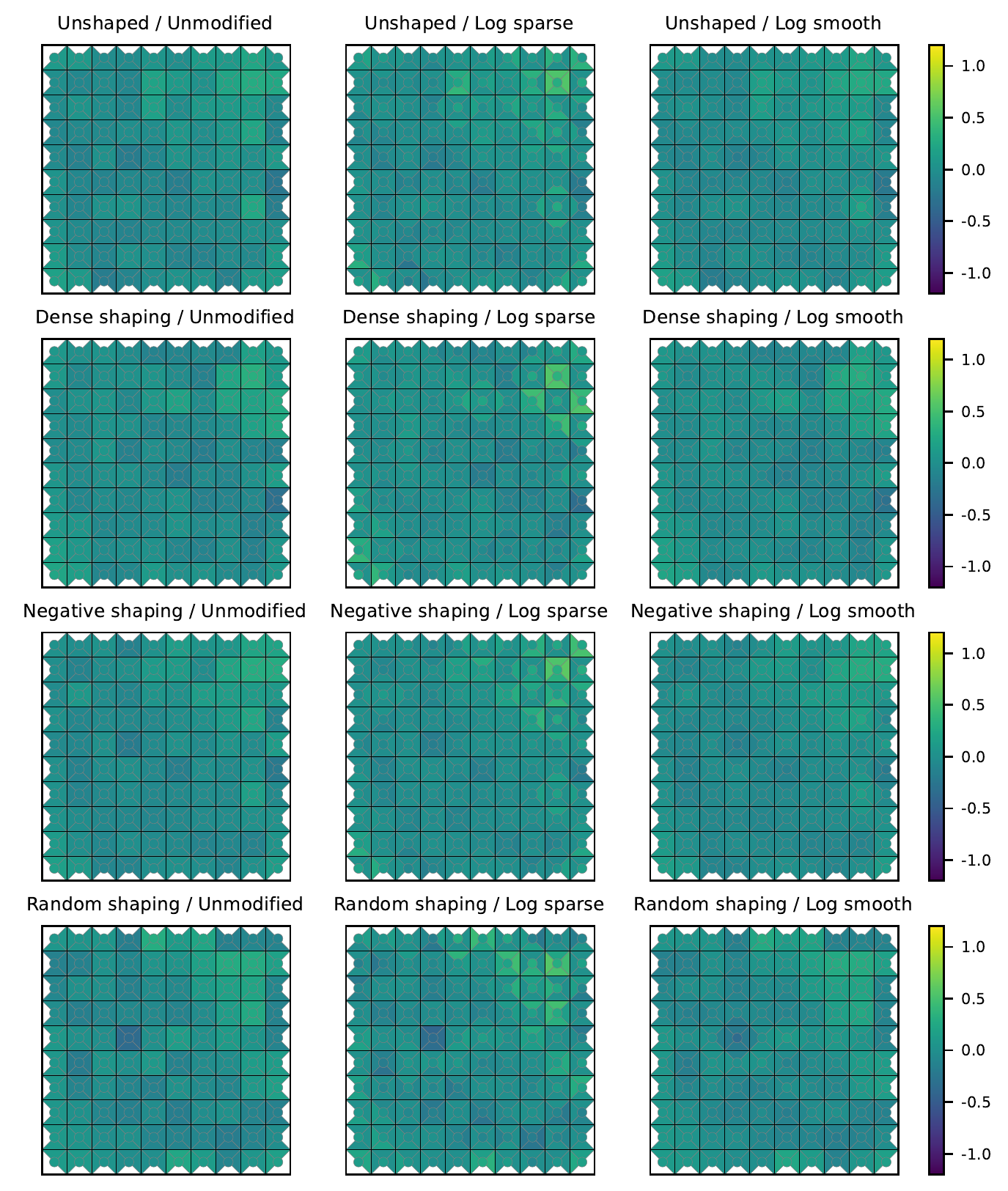}
  \caption{
    Reward models learned using AIRL from expert demonstrations for the \texttt{Goal} reward (leftmost column) and preprocessed versions of these (middle and right). The cost functions used here are the logarithmic versions of the sparsity and smoothness cost.
     \gridworldcaption{}
  }
  \label{fig:gridworld_airl_10_goal_log}
\end{figure}
\begin{figure}
  \centering
  \includegraphics[width=\textwidth]{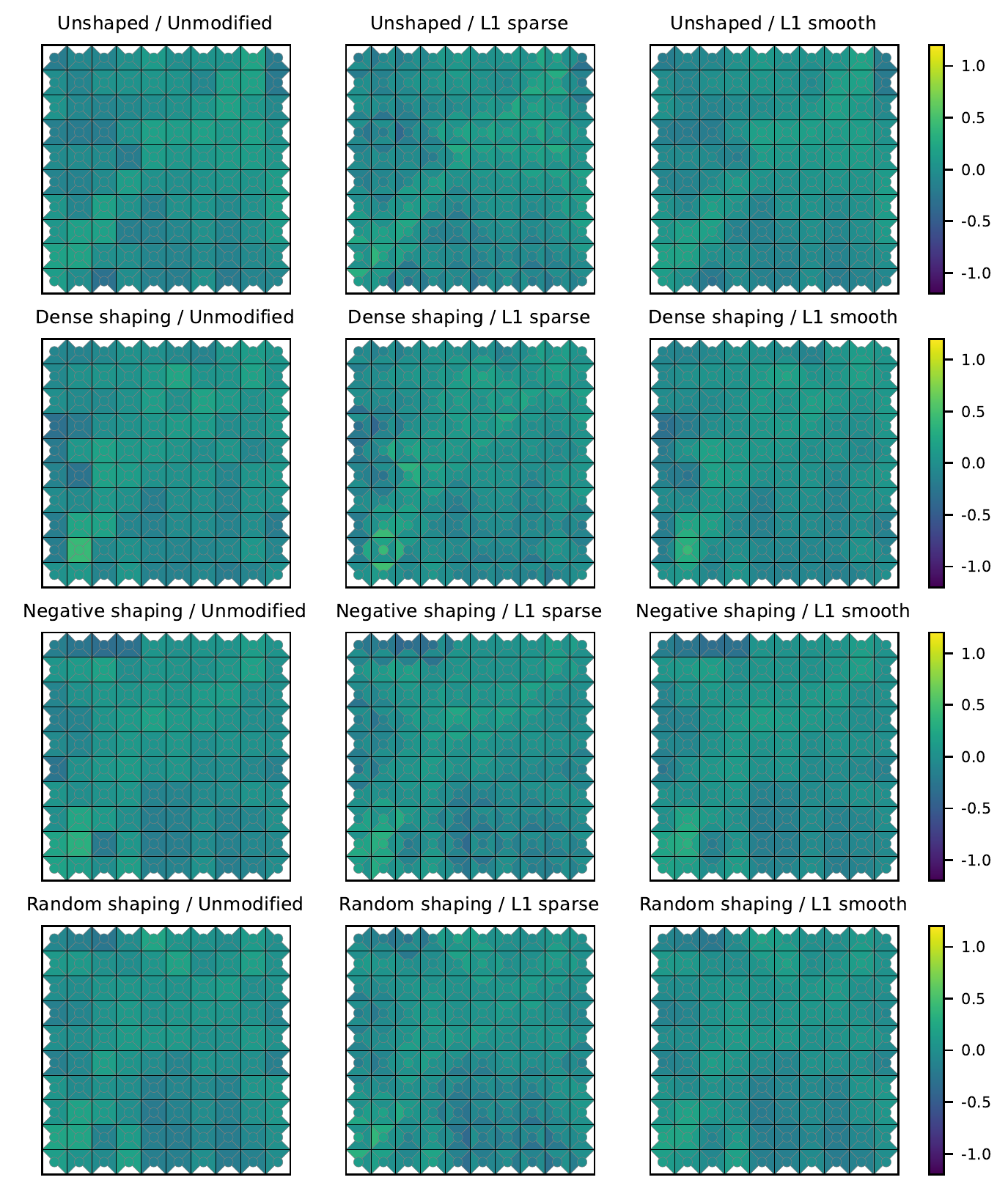}
  \caption{
    Reward models learned using AIRL from expert demonstrations for the \texttt{Path} reward (leftmost column) and preprocessed versions of these (middle and right). The cost functions used here are the \(L^1\) versions of the sparsity and smoothness cost.
     \gridworldcaption{}
  }
  \label{fig:gridworld_airl_10_path_l1}
\end{figure}
\begin{figure}
  \centering
  \includegraphics[width=\textwidth]{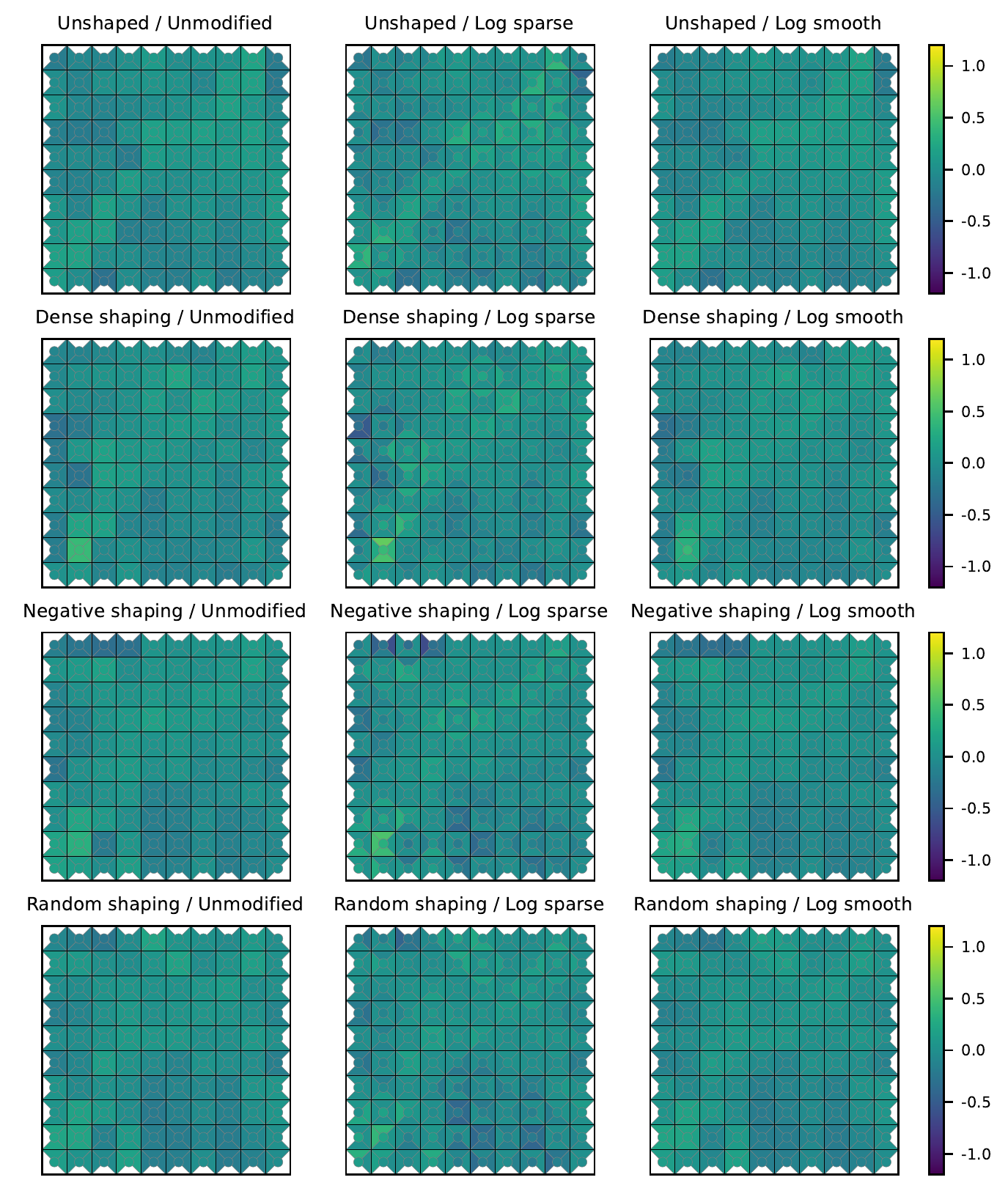}
  \caption{
    Reward models learned using AIRL from expert demonstrations for the \texttt{Path} reward (leftmost column) and preprocessed versions of these (middle and right). The cost functions used here are the logarithmic versions of the sparsity and smoothness cost.
     \gridworldcaption{}
  }
  \label{fig:gridworld_airl_10_path_log}
\end{figure}

\begin{figure}
  \centering
  \includegraphics[width=\textwidth]{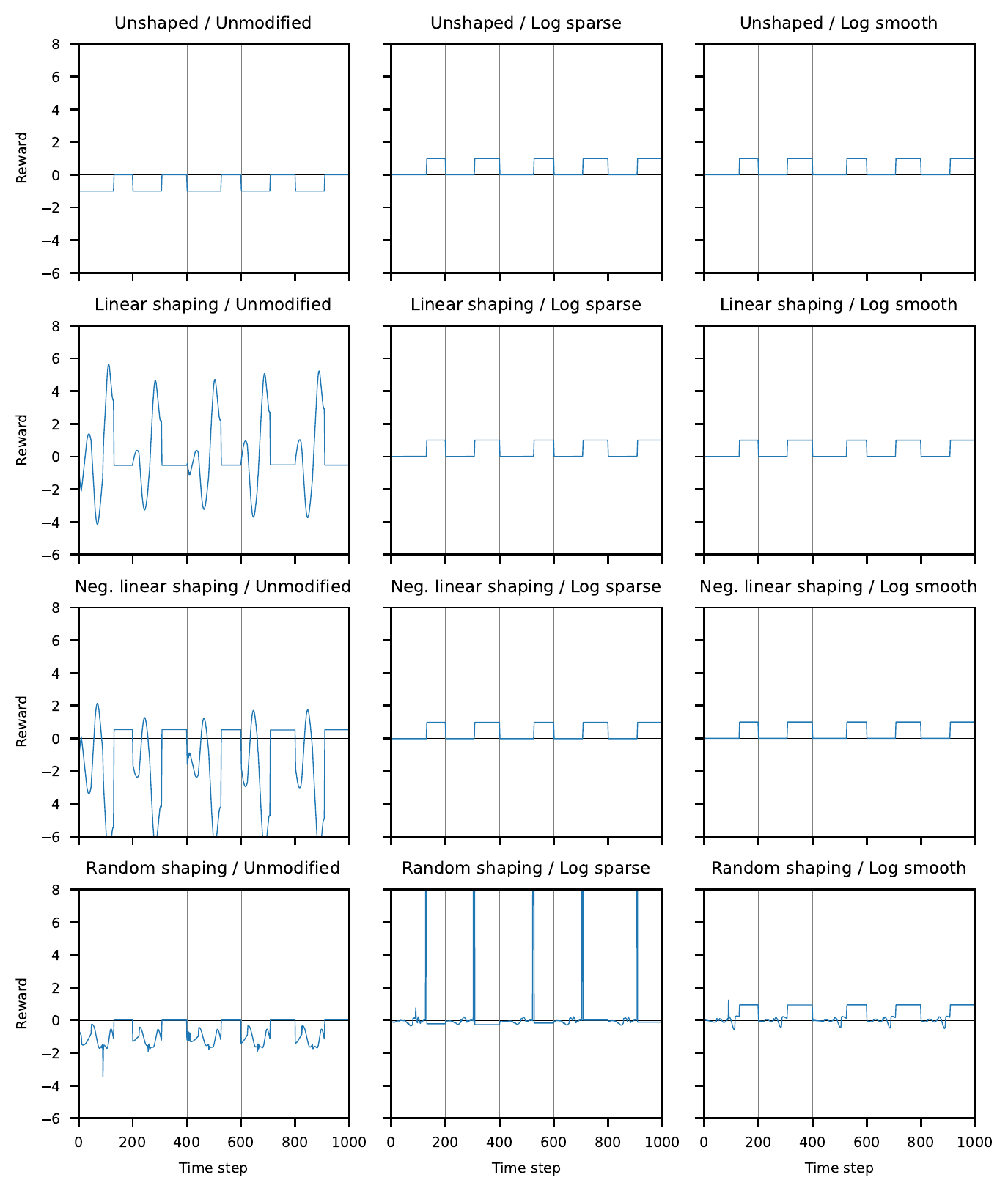}
  \caption{Preprocessing simplifies rewards in the continuous mountain car environment. The top-left shows the ground-truth reward over time, with three shaped versions below. The middle and right column show these rewards after preprocessing using the logarithmic sparsity and smoothness metrics. For the first two (linear) shapings, preprocessing recovers the ground truth reward exactly (up to a constant shift). In the more complex case in the last row, preprocessing still significantly simplifies the reward. See \cref{fig:mountain_car_ground_truth_l1} for versions with an \(L^1\) cost function. \mountaincarcaption{}}
  \label{fig:mountain_car_ground_truth}
\end{figure}

\begin{figure}
  \centering
  \includegraphics[width=\textwidth]{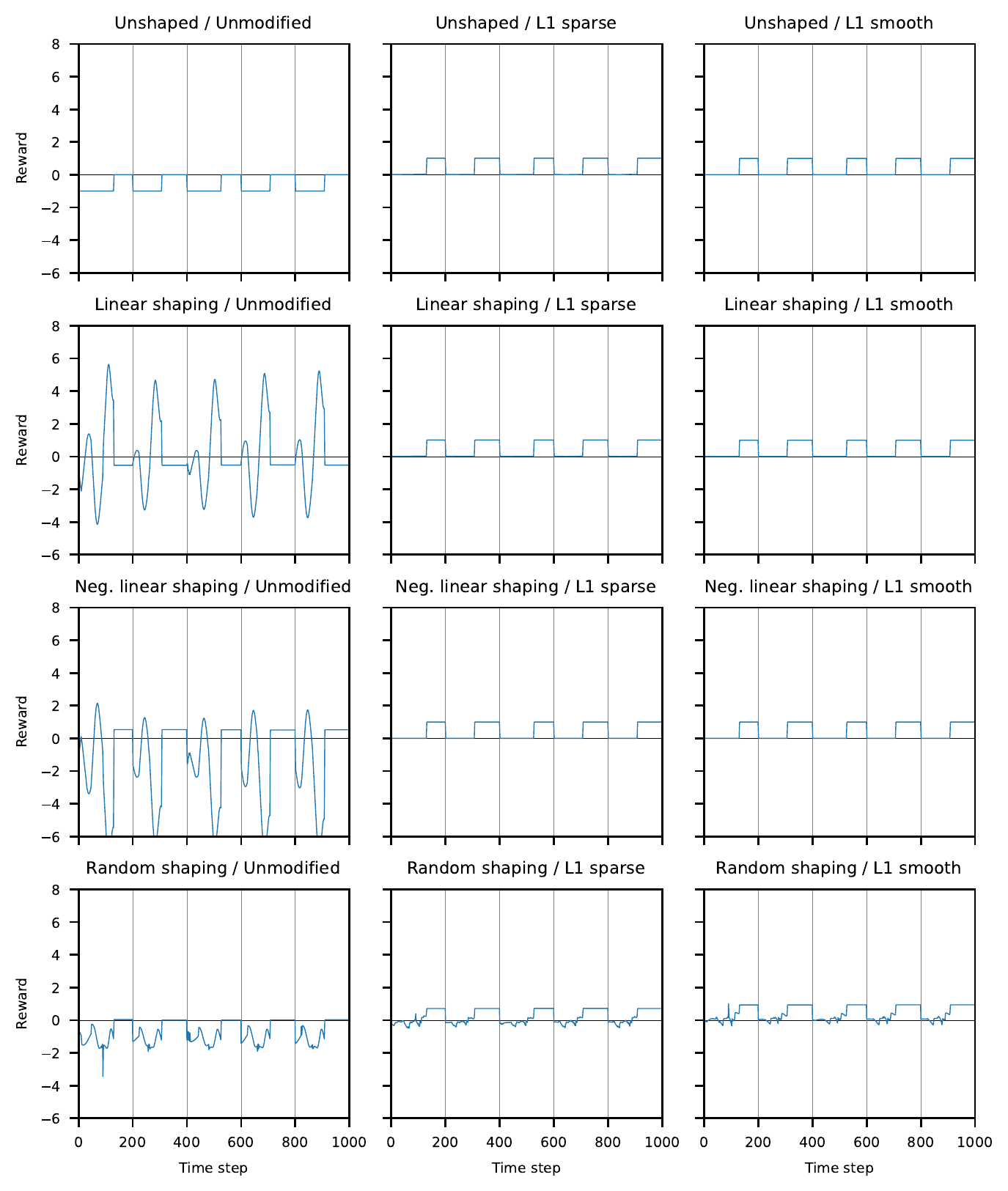}
  \caption{The top-left shows the ground-truth reward in mountain car over time, with three shaped versions below. The middle and right column show these rewards after preprocessing using the \(L^1\) sparsity and smoothness metrics. This works reasonably well for these simple shaped rewards, although in the more complex last row these cost functions appear to perform less well than the logarithmic version in \cref{fig:mountain_car_ground_truth}. \mountaincarcaption{}}
  \label{fig:mountain_car_ground_truth_l1}
\end{figure}

\begin{figure}
  \centering
  \includegraphics[width=\textwidth]{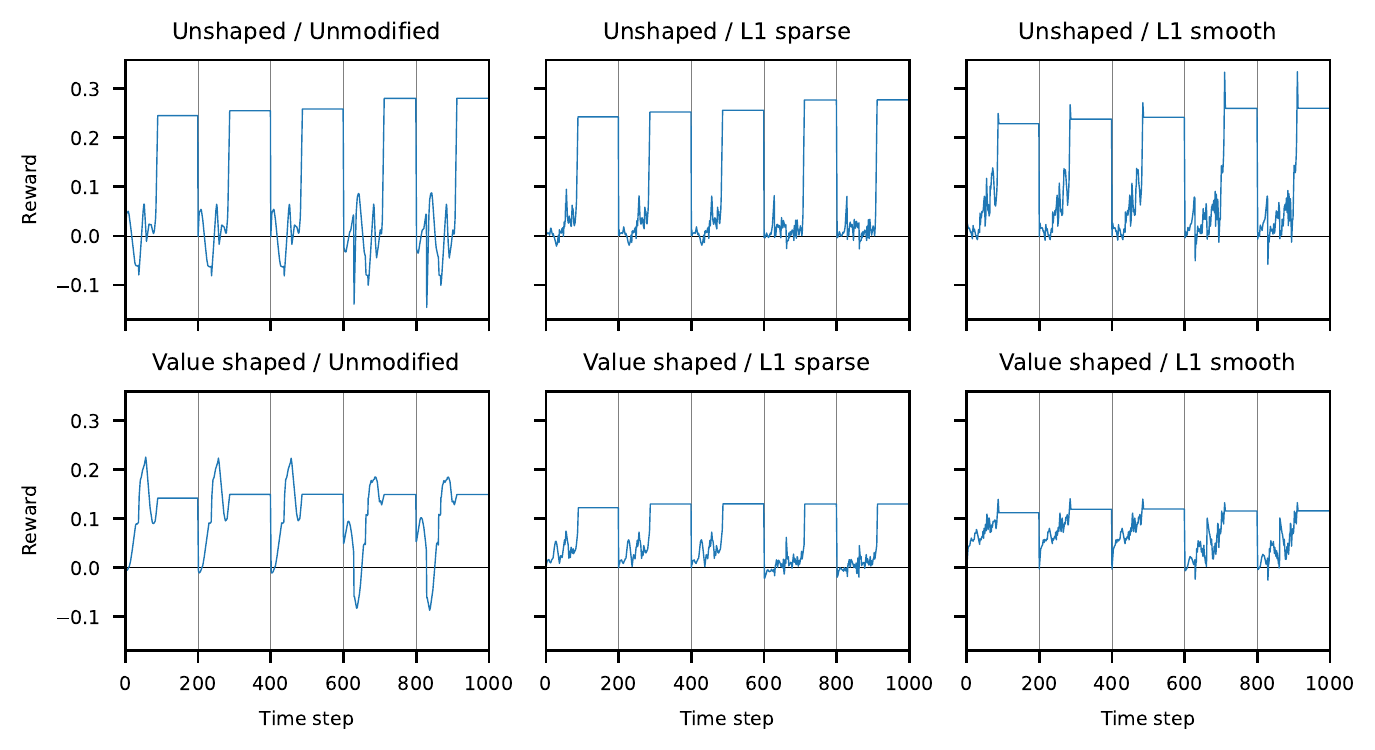}
  \caption{Preprocessing can simplify complex learned reward models for mountain car. The left column shows reward models learned using synthetic preference comparisons based on the ground-truth reward (top), and the ground-truth shaped with an optimal value function (bottom). Preprocessing for sparsity (middle) and smoothness (right) produces simpler and less noisy reward curves, especially in the shaped setting. The results are extremely similar, although perhaps slightly worse, than the logarithmic version used in \cref{fig:mountain_car_drlhp_log}. \mountaincarcaption{}}
  \label{fig:mountain_car_drlhp_l1}
\end{figure}

\end{document}